\documentclass[10pt,twocolumn,letterpaper]{article}

\usepackage{iccv}              
\usepackage{comment}
\usepackage{balance}
\usepackage{multirow}

\definecolor{iccvblue}{rgb}{0.21,0.49,0.74}
\usepackage[pagebackref,breaklinks,colorlinks,allcolors=iccvblue]{hyperref}

\def\paperID{12} 
\def\confName{ICCV}
\def\confYear{2025}

\title{A Multi-domain Image Translative Diffusion StyleGAN for Iris Presentation Attack Detection}

\author{Shivangi Yadav\\
Michigan State University\\
East Lansing, Michigan\\
{\tt\small yadavshi@msu.edu}
\and
Arun Ross\\
Michigan State University\\
East Lansing, Michigan\\
{\tt\small rossarun@msu.edu}
}

\begin{document}
\maketitle
\begin{abstract}
An iris biometric system can be compromised by presentation attacks (PAs) where artifacts such as artificial eyes, printed eye images, or cosmetic contact lenses are presented to the system. To counteract this, several presentation attack detection (PAD) methods have been developed. However, there is a scarcity of datasets for training and evaluating iris PAD techniques due to the implicit difficulties in constructing and imaging PAs. To address this, we introduce the Multi-domain Image Translative Diffusion StyleGAN (MID-StyleGAN), a new framework for generating synthetic ocular images that captures the PA and bonafide characteristics in multiple domains such as bonafide, printed eyes and cosmetic contact lens. MID-StyleGAN combines the strengths of diffusion models and generative adversarial networks (GANs) to produce realistic and diverse synthetic data. Our approach utilizes a multi-domain architecture that enables the translation between bonafide ocular images and different PA domains. The model employs an adaptive loss function tailored for ocular data to maintain domain consistency. Extensive experiments demonstrate that MID-StyleGAN outperforms existing methods in generating high-quality synthetic ocular images. The generated data was used to significantly enhance the performance of PAD systems, providing a scalable solution to the data scarcity problem in iris and ocular biometrics. For example, on the LivDet2020 dataset, the true detect rate at 1\% false detect rate improved from 93.41\% to 98.72\%, showcasing the impact of the proposed method.
\end{abstract}    
\section{Introduction}
\label{sec:intro}

Iris-based biometric systems are known for their reliability and contactless recognition of individuals \cite{jain2016}. However, as these systems become more widespread, they are increasingly targeted by presentation attacks (PAs), where attackers attempt to deceive the system using artifacts such as printed images, textured cosmetic contact lens, artificial eyes, etc. to impersonate another real individual, create a virtual identity, or obfuscate their own identity \cite{czajka2018}. Detecting such attacks is important for a secure iris recognition system, but is hampered by the limited availability of pertinent iris and ocular datasets. This lack of data makes it difficult to adequately train models to recognize the subtle differences between bonafide and PAs, particularly when considering the wide range of variations within and across different PAs of iris images (such as printed eyes and cosmetic contact lens). One solution to overcome this challenge is to augment the training data with synthetic data that exhibit realistic images in both the bonafide and PA domains. These synthetic datasets can help in training as well as evaluating PA detection algorithms, ensuring that they are robust against a wide range of attacks \cite{kohli2017,yadav2021,yadav2019}.

\begin{figure}
    \centering
    \includegraphics[width=0.6\linewidth]{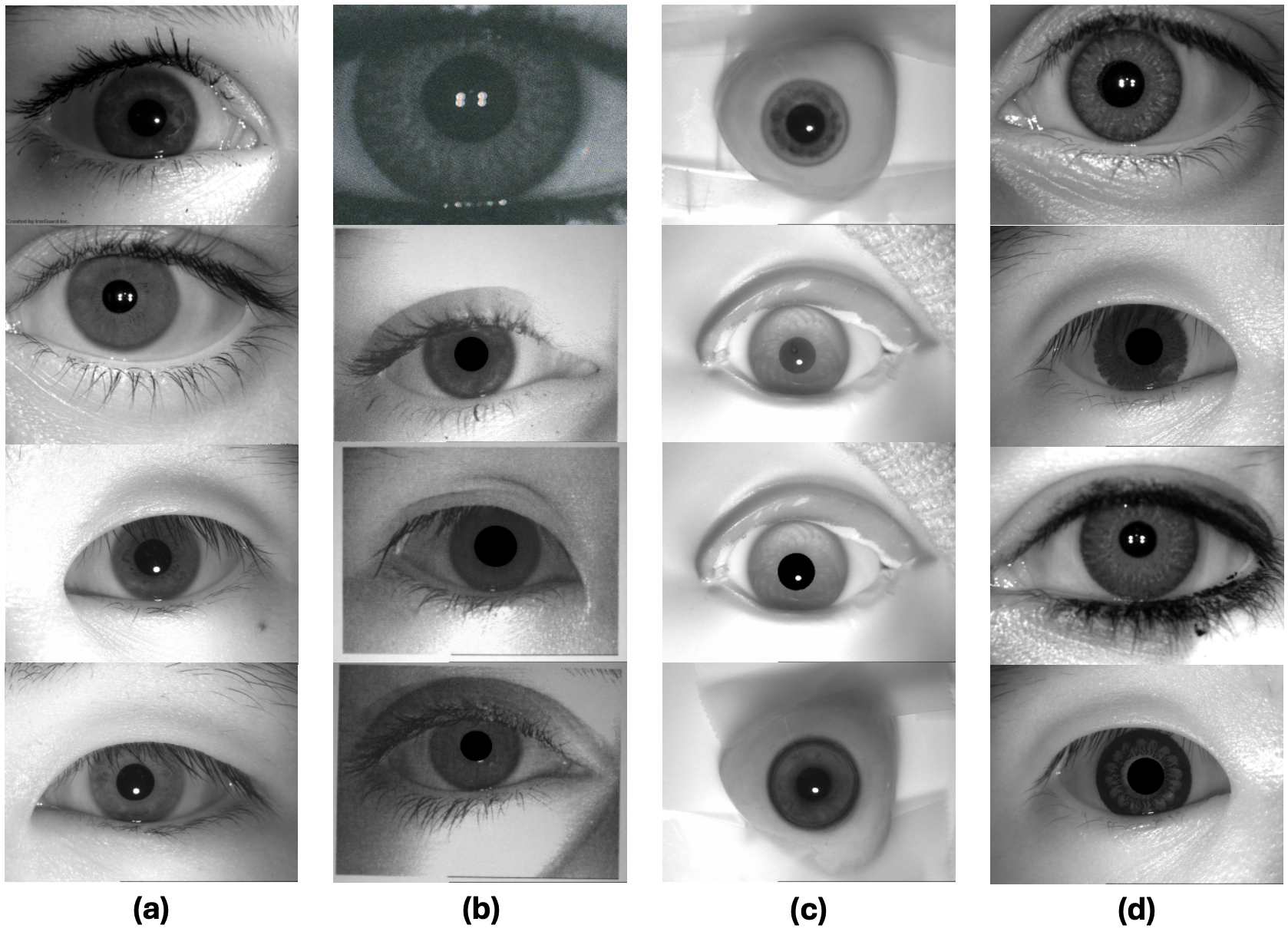}
    \caption{Examples of bonafide and different types of iris presentation attacks (PAs). (a) Bonafide, (b) printed eye, (c) artificial eye, and (d) cosmetic contact lens. These images are taken from Berc-IrisFake \cite{Lee2007}, CASIA-IrisFake \cite{Sun2014} and LivDet-2017 \cite{Yambay2017} datasets.}
    \label{fig:real-samples}
    \vspace{-3mm}
\end{figure}

The generation of realistic synthetic biometric data, including iris, has been explored in the literature. The more recent methods employ generative adversarial networks (GANs) to produce synthetic images \cite{yadav2019,patrick2022,wyzykowski2023, engelsma2022,han2018,fang2023}. GANs typically take a random noise vector as input and generate a realistic image from it. For instance, Kohli et al. \cite{kohli2017} proposed iDCGAN for synthesizing cropped iris images from random noise. While they showed good results for croppped iris images of size \ensuremath{64\times64}, this method struggles with generating higher resolution images, and fails in generating {\em ocular} images. Yadav et al. \cite{yadav2019} utilized the Relativistic Average Standard Generative Network (RaSGAN) to generate high-resolution iris images from random noise input. Another category of GANs, focused primarily on tasks such as image editing and domain specific style transfer, are image translative GANs that takes an image as an input and generate a synthetic image as per conditions specified for image translation. For example, Richardson et al. \cite{richardson2021} proposed pSp, a image translative StyleGAN that takes a face image as an input and generates a synthetic image with altered style attributes such as hair color and expressions while keeping intact the characteristics that defines the identity of the face in the given input image. In another work, Yadav et al. \cite{yadav2021} proposed CIT-GAN that utilizes paired training data to translate a source iris image into a synthetic image that incorporates the attributes from a target domain defined using a reference iris image. This process allows the generator to map across different domains, making it versatile for multiple applications. While these methods offer significant improvements over traditional methods for synthetic {\em iris} image generation \cite{cui2004,shah2006,zuo2007}, the quality of images degrade for {\em ocular} images where sometimes GANs focus too much on the non-iris parts of the images (such as eyelashes) while failing to capture the intricate details of the iris.

In this paper, we address the problem of generating realistic high resolution ocular images while overcoming the shortcomings of GANs (mode collapse, unstable training, etc.). Ocular images provide richer context and additional information compared to the cropped iris images. It includes not only the iris but also the surrounding regions such as the sclera, eyelashes, and eyelids. These elements play a key role in many biometric applications such as PA detection (PAD), where adversarial artifacts might appear beyond the iris itself. Additionally, generating ocular images facilitates the development of more robust machine learning models that can handle diverse real-world scenarios. To generate such images with rich contextual information we propose a novel approach, known as Multi-domain Image Translative Diffusion StyleGAN (MID-StyleGAN), to generate realistic high resolution ocular synthetic PA datasets. This method combines the strengths of StyleGAN \cite{karras2020, karras2021} and diffusion models \cite{wang2022,xiao2021} for high-fidelity ocular image synthesis while utilizing a multi-domain diffusion timestep-dependent discriminator and an image encoder for smooth transitions and variations across multiple PA domains, i.e., the discriminator is responsible for distinguishing between real and synthetic images, as well as classifying the domain of the image (e.g., determining whether the image belongs to the domain category of bonafide, printed eye or cosmetic contact lens). Also, the ocular image encoder utilizes feedback from the discriminator to learn domain-specific knowledge. This helps the network to better learn image translation from source to target domain. In Section \ref{sec:exp}, we will show how the images generated using the proposed method are not only more realistic than those from other GAN methods, but also capture the inter- and intra-domain variations (as shown in Figure \ref{fig:real-samples}). Further, we will show how the dataset of synthetic irides can be utilized for enhancing the performance of PA detection (PAD) methods.

The contributions of this paper are as follows:
\begin{itemize}
    \item We propose a Multi-domain Image Translative Diffusion StyleGAN (MID-StyleGAN) that combines the strengths of GANs and diffusion models to generate realistic high-resolution ocular synthetic PA and bonafide datasets.
    \item The proposed method (a) utilizes forward diffusion process in combination with GANs to generate high-resolution, realistic synthetic images, (b) employs a multi-domain diffusion timestep-dependent discriminator that is scalable to multiple domains, and (c) promotes domain transfer using conditional adversarial training and domain transfer loss.
    \item We compare and analyze the realism of ocular images generated by our proposed method with other methods in the literature.
    \item We evaluate the utility of the generated ocular PA dataset in enhancing the performance of a DNN-based PA detector.
\end{itemize}

\begin{figure*}
    \centering
    \includegraphics[width=0.65\linewidth]{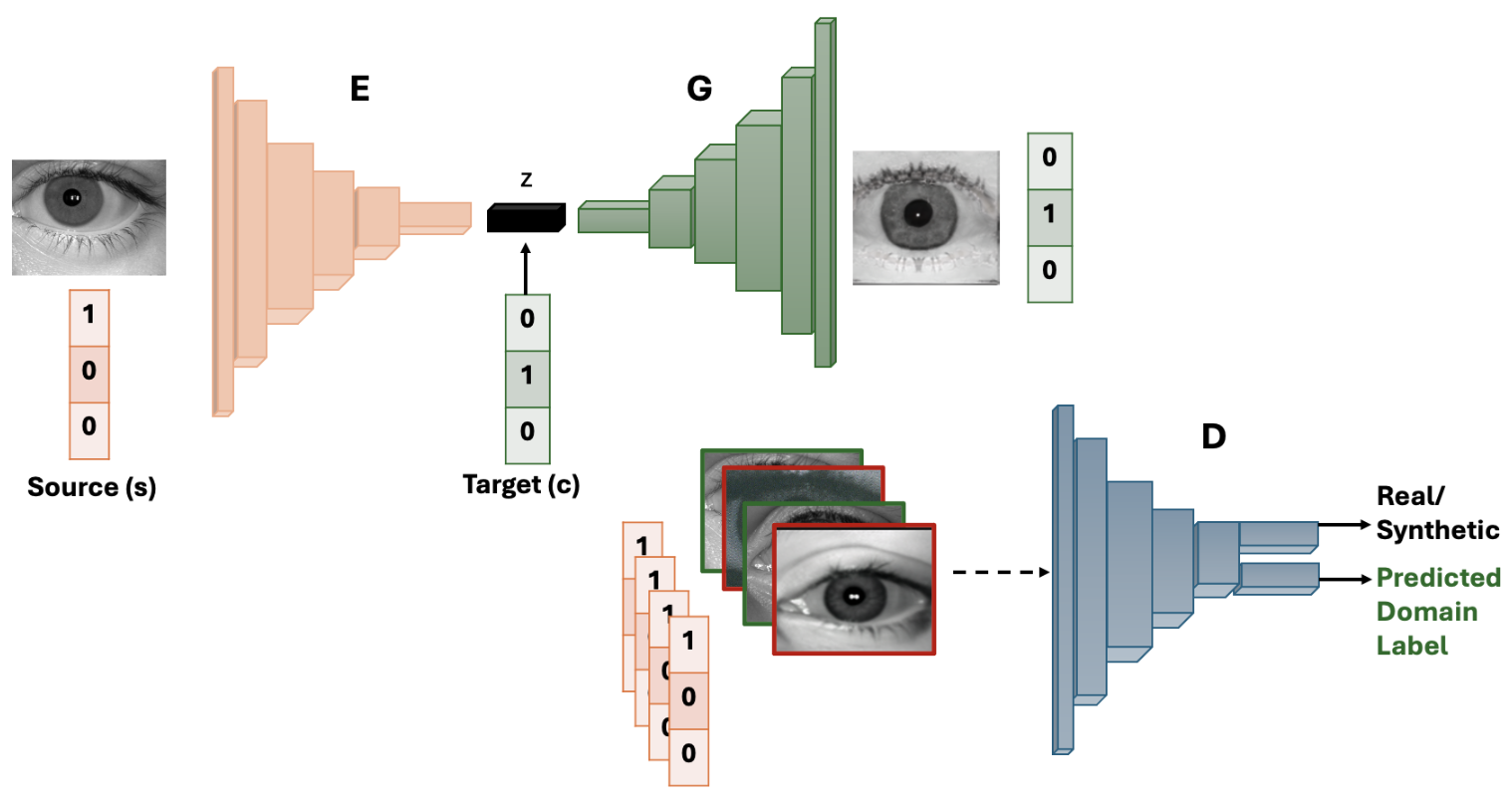}
    \caption{Illustration of the proposed method that has three modules: (1) Encoder, \ensuremath{E}, which takes an image and its domain label as an input and outputs the encoded image, (2) \ensuremath{G} that takes the encoded image as an input along with the target domain label to which the input image has to be translated, and (3) \ensuremath{D} that takes an image and its label as input, and outputs the image probability of domains as well as whether the image is real or synthetic.}
    \label{fig:arch}
\end{figure*}

\begin{figure}
    \centering
    \includegraphics[width=0.95\linewidth]{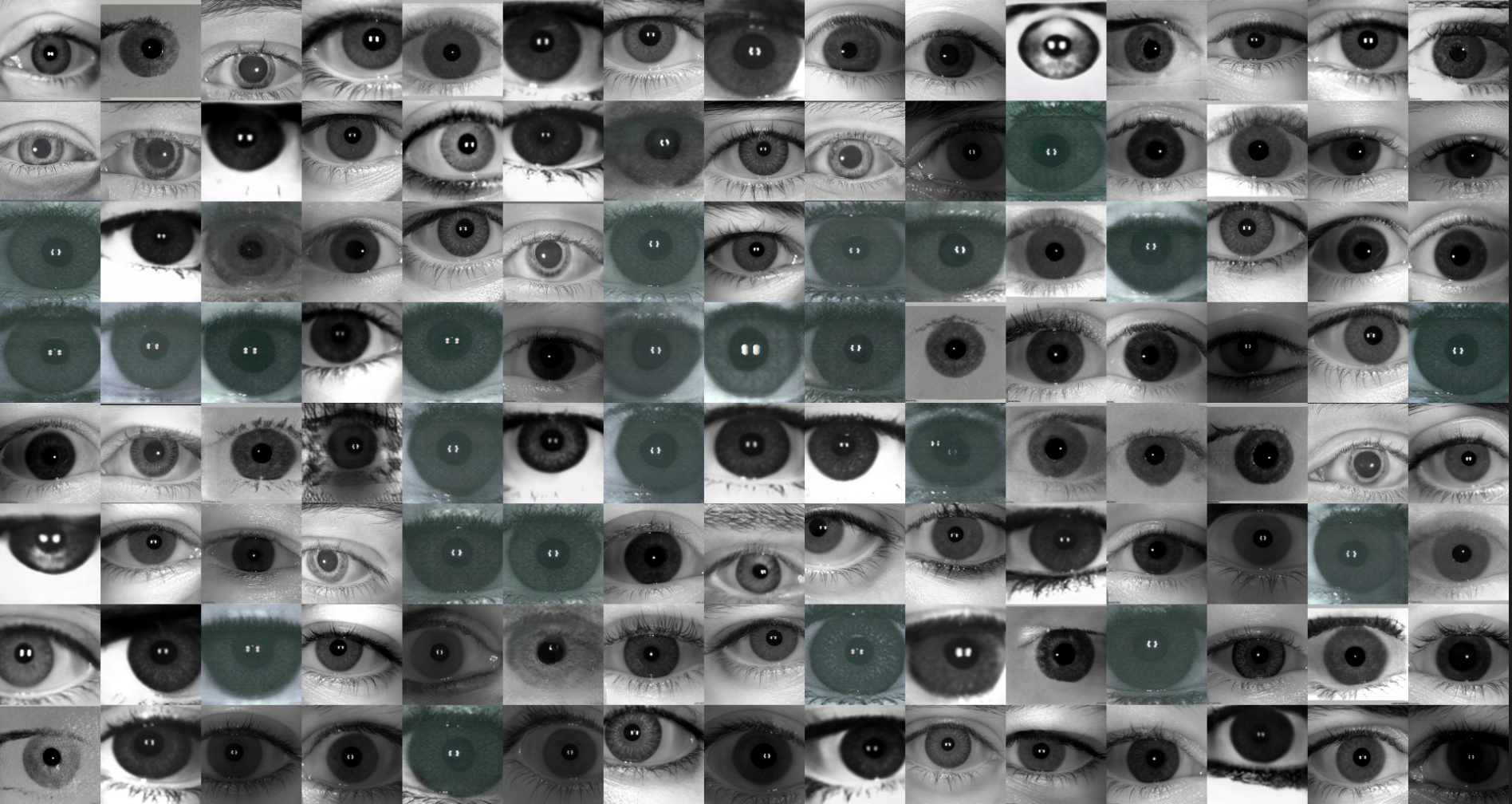}
    \caption{Samples of ocular images generated using proposed method, MID-StyleGAN. The proposed method is capable of not only generating images from multiple PA domains but also capture intra class variations present in different types of PAs.}
    \label{fig:synthetic}
    \vspace{-2mm}
\end{figure}

\section{Background}
\label{sec:background}

\subsection{Generative Adversarial Networks (GANs)}
Generative Adversarial Networks (GANs) \cite{goodfellow2020} are generative models that transform random noise into realistic synthetic images. A GAN consists of two components: (1) a Generator (\ensuremath{G}) that creates images and (2) a Discriminator (\ensuremath{D}) that distinguishes real from synthetic images. Trained adversarially in a min-max game, \ensuremath{G} aims to deceive \ensuremath{D}, while \ensuremath{D} learns to distinguish between real and synthetic images.

\subsection{Diffusion-GANs}
Researchers have made significant progress in improving the quality and realism of images generated by GANs. However, challenges such as artifacts, distortions, and mode collapse—where the generator fails to capture the full diversity of the target distribution—still persist. Diffusion Generative Adversarial Networks (Diffusion-GANs) \cite{wang2022} offer a promising approach to address some of these issues. In Diffusion-GANs, researchers introduced an innovative GAN framework that utilizes a forward diffusion process to generate Gaussian-mixture distributed instance noise, addressing some of the key challenges in GAN training such as instability and mode collapse. The Diffusion-GAN framework comprises three key components: an adaptive diffusion process, a diffusion timestep-dependent discriminator, and a generator. Both real and generated data are subjected to the same adaptive diffusion process, which progressively adjusts the noise-to-data ratio across different timesteps. This enables the model to refine the data transformation in a controlled manner.

Combining the strengths of GANs and diffusion models, Diffusion GANs leverage the diffusion process to guide the generator in producing realistic samples, while the adversarial component ensures that the generated data is indistinguishable from real data. To incorporate the diffusion process into generative adversarial training, the original min-max objective \cite{goodfellow2020} is modified as follows \cite{wang2022}:
\vspace{-3mm}

\begin{equation}
    \begin{split}
        \mathcal{L}(D,G) = \mathbb{E}_{x \sim p(x), t \sim p_\pi, y \sim q(y \mid x, t)}
        \left[\log(D(y, t))\right] \\
        + \mathbb{E}_{z \sim p(z), t \sim p_\pi, y_g \sim q(y \mid G_\theta(z), t)}
        \left[\log(1 - D(y_g, t))\right]
    \end{split}
    \label{eq:adv}
\end{equation}

Here, \ensuremath{p(x)} refers to the distribution of the real data and \ensuremath{p_\pi} refers to a discrete distribution that helps assign weights \ensuremath{\pi_t} to each step \ensuremath{t \in {1,....T}}, in the diffusion process. Also, \ensuremath{\boldsymbol{y}} and \ensuremath{\boldsymbol{y_g}}, respectively, refer to the noisy and generated noisy counterpart of the real image \ensuremath{x}. With the introduction of a diffusion process, Diffusion-GANs require a new optimization strategy for the discriminator (\ensuremath{D}) to effectively distinguish between real and synthetic images. This is achieved by having the discriminator learn from the simplest examples (with no noise) while gradually increasing the noise-to-data ratio. A self-paced schedule is used to determine the number of diffusion steps (\ensuremath{T}), based on a discriminator over-fitting metric (\ensuremath{r_d}). This metric, derived from \cite{karras2020}, evaluates the discriminator's confidence relative to the data:
\begin{equation}
    \begin{split}
        r_d &= \mathbb{E}_{\boldsymbol{x_n},t \sim p(\boldsymbol{x_n},t)} [\text{sign}(D(\boldsymbol{x_n}, t) - 0.5)], \\
T &= T + \text{sign}(r_d - d_{\text{target}}) \times C
    \end{split}
\end{equation}

The schedule adjusts \ensuremath{T} based on the deviation of \ensuremath{r_d} from a target value, with a constant factor (C) influencing the rate of change. \ensuremath{r_d} is recalculated and \ensuremath{T} updated after every four mini-batches \cite{karras2020}.

\section{Proposed Method}
\label{sec:propose}
The proposed Multi-domain Image Translative Diffusion StyleGAN (MID-StyleGAN) model is designed to generate synthetic ocular images that capture the diversity found in real-world datasets (as shown in Figure \ref{fig:real-samples}). This method utilizes StyleGAN-3 as the backbone architecture while incorporating a forward diffusion process to generate high-resolution, realistic ocular images. Specifically, both real and generated data undergo an adaptive diffusion process, which dynamically adjusts the noise-to-data ratio across timesteps.

A key component of the framework is the diffusion timestep-dependent multi-domain discriminator \ensuremath{D}, which evolves with the diffusion process and is tasked with distinguishing diffused real data from diffused generated data at each timestep. The generator benefits from this forward diffusion chain, which adaptively adjusts its length to maintain an optimal balance between noise and data levels. This interaction stabilizes the adversarial training process, reduces mode collapse, and ensures high-fidelity image synthesis. The discriminator in the proposed MID-StyleGAN is not only tasked to distinguish between real and synthetic images, but also classify the domain of the image (e.g., determining whether the image is a bonafide or one of many PA types). Also, the image encoder, \ensuremath{E}, learns to encode an ocular image while utilizing feedback from the discriminator to learn domain-specific knowledge. This helps the network to learn a smooth image translation from the source to the target domain. To achieve this, the adversarial loss in Eqn. (\ref{eq:adv}) has to be modified as,
\vspace{-2mm}

\begin{equation}
    \begin{split}
        \mathcal{L}(D,G) = \mathbb{E}_{x \sim p(x), t \sim p_\pi, y \sim q(y \mid x, t)}
        \left[\log(D(y, t))\right] \\
        + \mathbb{E}_{t \sim p_\pi, y_g \sim q(y \mid G_\theta(E(x,s),c), t)}
        \left[\log(1 - D(y_g, t))\right]
    \end{split}
    \label{eq:new_adv}
\end{equation}

Here, \ensuremath{c} refers to the target domain and \ensuremath{E(.)} refers to the image encoder \ensuremath{E} that takes an image \ensuremath{x} as input and its domain label \ensuremath{s}. \ensuremath{\boldsymbol{y}} and \ensuremath{\boldsymbol{y_g}} refer to real noisy and generated noisy images, at step \ensuremath{t}. The sub-network in \ensuremath{D} for domain classification helps promote domain transfer using the following loss functions,

\begin{equation}
\mathcal{L}_{\text{domain}}^{\text{real}} = \mathbb{E}_{y \sim q(y \mid x, t)} \left[\log D_{\text{domain}}(s | y,t) \right]
\end{equation}

\begin{equation}
\mathcal{L}_{\text{domain}}^{\text{synthetic}} = \mathbb{E}_{y_g \sim q(y \mid G_\theta(z), t)} \left[ \log D_{\text{domain}}(c | y_g,t) \right]
\end{equation}

\begin{equation}
\mathcal{L}_{\text{domain}} = \mathcal{L}_{\text{domain}}^{\text{real}} + \mathcal{L}_{\text{domain}}^{\text{synthetic}}
\end{equation}

Here, \ensuremath{D_{domain}} represents the domain classifier component of the discriminator. In order to ensure that the encoder learns to translate input image \ensuremath{x} to latent code \ensuremath{z} that represents the iris and PA distributions, we define the content preservation loss as,
\vspace{-3mm}


\begin{equation}
    \mathcal{L}_{\text{recon}} = \mathbb{E}_{x \sim p_{\text{data}}(x)} \left[ \| G(E(x,s),c) - x \|_2^2 \right]
\end{equation}

\begin{equation}
    \mathcal{L}_{\text{LPIPS}} = \mathbb{E}_{x \sim p_{\text{data}}(x)} \left[ \sum_{l} \| \phi_l(G(E(x,s),c)) - \phi_l(x) \|_2^2 \right]
\end{equation}

Here, \ensuremath{\phi_l(.)} represents the features extracted by layer \ensuremath{l} of a pre-trained network.\footnote{\url{https://github.com/TreB1eN/InsightFace_Pytorch}} In order to ensure that the network does not alter the image drastically when it is already in the target domain and capturing the intra-class variations, we define the following loss:

\begin{equation}
    \mathcal{L}_{\text{inr}} = \mathbb{E}_{x \sim p_{\text{data}}(x)} \left[ \| G(E(x,c),c) - x \|_2^2 \right]
\end{equation}

Further, as described in \cite{karras2020}, to ensure diversity and encourage consistent image quality across domains, we employ a style-mixing technique. The objective is to prevent the generator from becoming too reliant on a single latent vector for generating images:
\vspace{-4mm}

\begin{equation}
    \mathcal{L}_{\text{mix}} = \sum_{i=1}^{L} \left\| G(E(x_1, s_1),c) - G(E(x_2, s_2), c) \right\|_2^2
\end{equation}

Here, \ensuremath{L} is the number of layers in the generator. 
\section{Experiments and Analysis}
\label{sec:exp}

\begin{table*}[h]
\centering
\caption{True Detection Rate (TDR) at different False Detection Rates (FDRs) for Baseline Experiment-0 when PA detectors are trained on D3: LivDet-2017 and tested using test sets from D1: Berc-iris-fake and D4: LivDet-2020. }
\scalebox{0.75}{
\begin{tabular}{|c|c|l|l|l|l|l|}
\hline
\textbf{Dataset} &                        & \multicolumn{1}{c|}{\textbf{VGG-19} \cite{simonyan2014}} & \multicolumn{1}{c|}{\textbf{AlexNet} \cite{krizhevsky2012}} & \multicolumn{1}{c|}{\textbf{ResNet-101} \cite{he2016}} & \multicolumn{1}{c|}{\textbf{MobileNet-v2} \cite{sandler2018}} & \multicolumn{1}{c|}{\textbf{DNetPAD} \cite{sharma20}} \\ \hline

\multirow{3}{*}{\textbf{D1}} 
& \textbf{TDR @ 1\% FDR} 
& \multicolumn{1}{c|}{92.12} & \multicolumn{1}{c|}{93.77} & \multicolumn{1}{c|}{99.08} & \multicolumn{1}{c|}{97.44} & \multicolumn{1}{c|}{93.41} \\ \cline{2-7}
& \textbf{TDR @ 2\% FDR} 
& \multicolumn{1}{c|}{97.99} & \multicolumn{1}{c|}{95.97} & \multicolumn{1}{c|}{99.45} & \multicolumn{1}{c|}{97.99} & \multicolumn{1}{c|}{95.60} \\ \cline{2-7}
& \textbf{TDR @ 5\% FDR} 
& \multicolumn{1}{c|}{99.08} & \multicolumn{1}{c|}{97.99} & \multicolumn{1}{c|}{99.63} & \multicolumn{1}{c|}{98.90} & \multicolumn{1}{c|}{99.08} \\ \hline

\multirow{3}{*}{\textbf{D4}} 
& \textbf{TDR @ 1\% FDR} 
& \multicolumn{1}{c|}{70.63} & \multicolumn{1}{c|}{48.96} & \multicolumn{1}{c|}{75.73} & \multicolumn{1}{c|}{76.51} & \multicolumn{1}{c|}{70.15} \\ \cline{2-7}
& \textbf{TDR @ 2\% FDR} 
& \multicolumn{1}{c|}{74.86} & \multicolumn{1}{c|}{54.47} & \multicolumn{1}{c|}{80.66} & \multicolumn{1}{c|}{79.96} & \multicolumn{1}{c|}{75.16} \\ \cline{2-7}
& \textbf{TDR @ 5\% FDR} 
& \multicolumn{1}{c|}{80.86} & \multicolumn{1}{c|}{61.69} & \multicolumn{1}{c|}{88.30} & \multicolumn{1}{c|}{84.40} & \multicolumn{1}{c|}{82.05} \\ \hline
\end{tabular}}
\label{tab:pa_comparison3}
\end{table*}

\begin{table*}[h]
\centering
\caption{True Detection Rate (TDR) at different False Detection Rates (FDRs) for Utility Experiment-1 when PA detectors are trained on D3: LivDet-2017 + Synthetic images and tested using test sets from D1: Berc-iris-fake and D4: LivDet-2020.}
\scalebox{0.75}{
\begin{tabular}{|c|c|l|l|l|l|l|}
\hline
\textbf{Dataset} &                        & \multicolumn{1}{c|}{\textbf{VGG-19} \cite{simonyan2014}} & \multicolumn{1}{c|}{\textbf{AlexNet} \cite{krizhevsky2012}} & \multicolumn{1}{c|}{\textbf{ResNet-101} \cite{he2016}} & \multicolumn{1}{c|}{\textbf{MobileNet-v2} \cite{sandler2018}} & \multicolumn{1}{c|}{\textbf{DNetPAD} \cite{sharma20}} \\ \hline

\multirow{3}{*}{\textbf{D1}} 
& \textbf{TDR @ 1\% FDR} 
& \multicolumn{1}{c|}{94.51} & \multicolumn{1}{c|}{97.99} & \multicolumn{1}{c|}{99.27} & \multicolumn{1}{c|}{98.90} & \multicolumn{1}{c|}{98.72} \\ \cline{2-7}
& \textbf{TDR @ 2\% FDR} 
& \multicolumn{1}{c|}{97.99} & \multicolumn{1}{c|}{97.99} & \multicolumn{1}{c|}{99.45} & \multicolumn{1}{c|}{99.08} & \multicolumn{1}{c|}{98.72} \\ \cline{2-7}
& \textbf{TDR @ 5\% FDR} 
& \multicolumn{1}{c|}{100}   & \multicolumn{1}{c|}{98.90} & \multicolumn{1}{c|}{99.82} & \multicolumn{1}{c|}{99.63} & \multicolumn{1}{c|}{99.72} \\ \hline

\multirow{3}{*}{\textbf{D4}} 
& \textbf{TDR @ 1\% FDR} 
& \multicolumn{1}{c|}{73.08} & \multicolumn{1}{c|}{54.60} & \multicolumn{1}{c|}{83.20} & \multicolumn{1}{c|}{79.39} & \multicolumn{1}{c|}{70.88} \\ \cline{2-7}
& \textbf{TDR @ 2\% FDR} 
& \multicolumn{1}{c|}{77.36} & \multicolumn{1}{c|}{60.91} & \multicolumn{1}{c|}{88.66} & \multicolumn{1}{c|}{83.39} & \multicolumn{1}{c|}{76.21} \\ \cline{2-7}
& \textbf{TDR @ 5\% FDR} 
& \multicolumn{1}{c|}{84.07} & \multicolumn{1}{c|}{68.89} & \multicolumn{1}{c|}{92.73} & \multicolumn{1}{c|}{86.97} & \multicolumn{1}{c|}{83.05} \\ \hline
\end{tabular}}
\label{tab:pa_comparison4}
\end{table*}

\begin{table*}[h]
\centering
\caption{True Detection Rate (TDR) at different False Detection Rates (FDRs) for Baseline Experiment-0 when PA detectors are trained on D1: Berc-Iris-Fake and tested using test sets from D3: LivDet-2017 and D4: LivDet-2020.}
\scalebox{0.75}{
\begin{tabular}{|c|c|l|l|l|l|l|}
\hline
\textbf{Dataset} &                        & \multicolumn{1}{c|}{\textbf{VGG-19} \cite{simonyan2014}} & \multicolumn{1}{c|}{\textbf{AlexNet} \cite{krizhevsky2012}} & \multicolumn{1}{c|}{\textbf{ResNet-101} \cite{he2016}} & \multicolumn{1}{c|}{\textbf{MobileNet-v2} \cite{sandler2018}} & \multicolumn{1}{c|}{\textbf{DNetPAD} \cite{sharma20}} \\ \hline

\multirow{3}{*}{\textbf{D3}} 
& \textbf{TDR @ 1\% FDR} 
& \multicolumn{1}{c|}{38.04} & \multicolumn{1}{c|}{24.07} & \multicolumn{1}{c|}{44.60} & \multicolumn{1}{c|}{55.70} & \multicolumn{1}{c|}{51.33} \\ \cline{2-7}
& \textbf{TDR @ 2\% FDR} 
& \multicolumn{1}{c|}{41.78} & \multicolumn{1}{c|}{28.24} & \multicolumn{1}{c|}{47.82} & \multicolumn{1}{c|}{58.69} & \multicolumn{1}{c|}{53.60} \\ \cline{2-7}
& \textbf{TDR @ 5\% FDR} 
& \multicolumn{1}{c|}{49.59} & \multicolumn{1}{c|}{35.44} & \multicolumn{1}{c|}{53.78} & \multicolumn{1}{c|}{65.28} & \multicolumn{1}{c|}{58.12} \\ \hline

\multirow{3}{*}{\textbf{D4}} 
& \textbf{TDR @ 1\% FDR} 
& \multicolumn{1}{c|}{27.40} & \multicolumn{1}{c|}{17.86} & \multicolumn{1}{c|}{34.26} & \multicolumn{1}{c|}{20.79} & \multicolumn{1}{c|}{21.03} \\ \cline{2-7}
& \textbf{TDR @ 2\% FDR} 
& \multicolumn{1}{c|}{38.72} & \multicolumn{1}{c|}{24.90} & \multicolumn{1}{c|}{45.60} & \multicolumn{1}{c|}{24.72} & \multicolumn{1}{c|}{25.36} \\ \cline{2-7}
& \textbf{TDR @ 5\% FDR} 
& \multicolumn{1}{c|}{55.88} & \multicolumn{1}{c|}{38.22} & \multicolumn{1}{c|}{61.54} & \multicolumn{1}{c|}{29.58} & \multicolumn{1}{c|}{32.89} \\ \hline
\end{tabular}}
\label{tab:pa_comparison1}
\end{table*}

\begin{table*}[h]
\centering
\caption{True Detection Rate (TDR) at different False Detection Rates (FDRs) for Utility Experiment-1 when PA detectors are trained on D1: Berc-Iris-Fake + Synthetic images and tested using test sets from D3: LivDet-2017 and D4: LivDet-2020.}
\scalebox{0.75}{
\begin{tabular}{|c|c|l|l|l|l|l|}
\hline
\textbf{Dataset} &                        & \multicolumn{1}{c|}{\textbf{VGG-19} \cite{simonyan2014}} & \multicolumn{1}{c|}{\textbf{AlexNet} \cite{krizhevsky2012}} & \multicolumn{1}{c|}{\textbf{ResNet-101} \cite{he2016}} & \multicolumn{1}{c|}{\textbf{MobileNet-v2} \cite{sandler2018}} & \multicolumn{1}{c|}{\textbf{DNetPAD} \cite{sharma20}} \\ \hline

\multirow{3}{*}{\textbf{D3}} 
& \textbf{TDR @ 1\% FDR}
& \multicolumn{1}{c|}{43.76} & \multicolumn{1}{c|}{28.99} & \multicolumn{1}{c|}{51.85} & \multicolumn{1}{c|}{57.11} & \multicolumn{1}{c|}{54.36} \\ \cline{2-7}
& \textbf{TDR @ 2\% FDR} 
& \multicolumn{1}{c|}{49.19} & \multicolumn{1}{c|}{34.67} & \multicolumn{1}{c|}{56.02} & \multicolumn{1}{c|}{60.71} & \multicolumn{1}{c|}{57.87} \\ \cline{2-7}
& \textbf{TDR @ 5\% FDR} 
& \multicolumn{1}{c|}{58.52} & \multicolumn{1}{c|}{43.73} & \multicolumn{1}{c|}{65.29} & \multicolumn{1}{c|}{68.57} & \multicolumn{1}{c|}{65.12} \\ \hline

\multirow{3}{*}{\textbf{D4}} 
& \textbf{TDR @ 1\% FDR} 
& \multicolumn{1}{c|}{14.99} & \multicolumn{1}{c|}{18.91} & \multicolumn{1}{c|}{42.07} & \multicolumn{1}{c|}{24.75} & \multicolumn{1}{c|}{52.99} \\ \cline{2-7}
& \textbf{TDR @ 2\% FDR} 
& \multicolumn{1}{c|}{43.67} & \multicolumn{1}{c|}{26.24} & \multicolumn{1}{c|}{53.14} & \multicolumn{1}{c|}{33.19} & \multicolumn{1}{c|}{64.37} \\ \cline{2-7}
& \textbf{TDR @ 5\% FDR} 
& \multicolumn{1}{c|}{60.51} & \multicolumn{1}{c|}{39.74} & \multicolumn{1}{c|}{68.35} & \multicolumn{1}{c|}{50.36} & \multicolumn{1}{c|}{75.73} \\ \hline
\end{tabular}}
\label{tab:pa_comparison2}
\end{table*}

\subsection{Datasets and PA Detection Methods}
In this research, we utilized three different PA datasets, viz., D1: Berc-iris-fake \cite{Lee2007}, D2: Casia-iris-fake \cite{Sun2014}, D3: LivDet-2017 \cite{Yambay2017} and D4: test set of LivDet-2020 \cite{das2020}\footnote{LivDet-2020 does not have a training set.} for training and testing different iris presentation attack detection (PAD) algorithms. These ocular PA datasets contain bonafide images and images from different PA classes such as cosmetic contact lens, printed eye and artificial eye (as shown in Figure \ref{fig:real-samples}). Each dataset is divided into train and test set using a 70-30 split on each domain (bonafide, printed eye and cosmetic contact lens).

The proposed generative network, MID-StyleGAN, is trained using the train set of LivDet-2017 dataset containing bonafide, printed eye and cosmetic contact lens (three domains). Using the trained network, we generate 10,000 synthetic ocular images per domain. We evaluate these images for realism and utility in the sub-sections below.

\subsection{Realism Assessment}
With the rise of DeepFake technology, researchers have explored methods to assess synthetic data quality. Salimans et al. \cite{salimans2018} introduced the Inception Score, using a pre-trained Inception-V3 model to compare marginal and conditional label distributions, with higher scores indicating better quality. However, this method does not account for the real data distribution in its calculations. To address this, Heusel et al. \cite{heusel2017} proposed the Fréchet Inception Distance (FID), which compares the statistical distributions of real and synthetic data:
\vspace{-3mm}

\begin{equation}\label{eq2} \begin{split} \ensuremath{FID = \left | \boldsymbol{\mu_r} - \boldsymbol{\mu_s} \right |^2 + Tr(\boldsymbol{\Sigma_r} + \boldsymbol{\Sigma_s} - 2 \sqrt{\boldsymbol{\Sigma_r}\boldsymbol{\Sigma_s}})} \end{split} \end{equation}

In this equation, \ensuremath{\boldsymbol{\mu_s}, \boldsymbol{\mu_r}, \boldsymbol{\Sigma_s}}, and \ensuremath{\boldsymbol{\Sigma_r}} represent the statistics of the synthetic (\ensuremath{s}) and real (\ensuremath{r}) distributions. Since FID measures the distance between these two distributions, a lower FID score indicates better quality of the generated data. 
\vspace{2mm}

\noindent
As described earlier, for this experiment we train MID-StyleGAN with the train set of LivDet-2017 dataset and generate 10,000 images for each domain (bonafide, printed eyes and cosmetic contact lens) using test images from D1, D2, D3 and D4 as source images. For each of the generated images, their realism score is calculated against the distribution of real images (source) using FID. For the comparative study, we utilize CIT-GAN \cite{yadav2021}, StyleGAN-3 \cite{karras2021} and diffusion based StyleGAN-3 (diff-Style3) \cite{wang2022}.

\begin{figure}
    \centering
    \begin{subfigure}[b]{0.45\textwidth}
        \includegraphics[width=\textwidth]{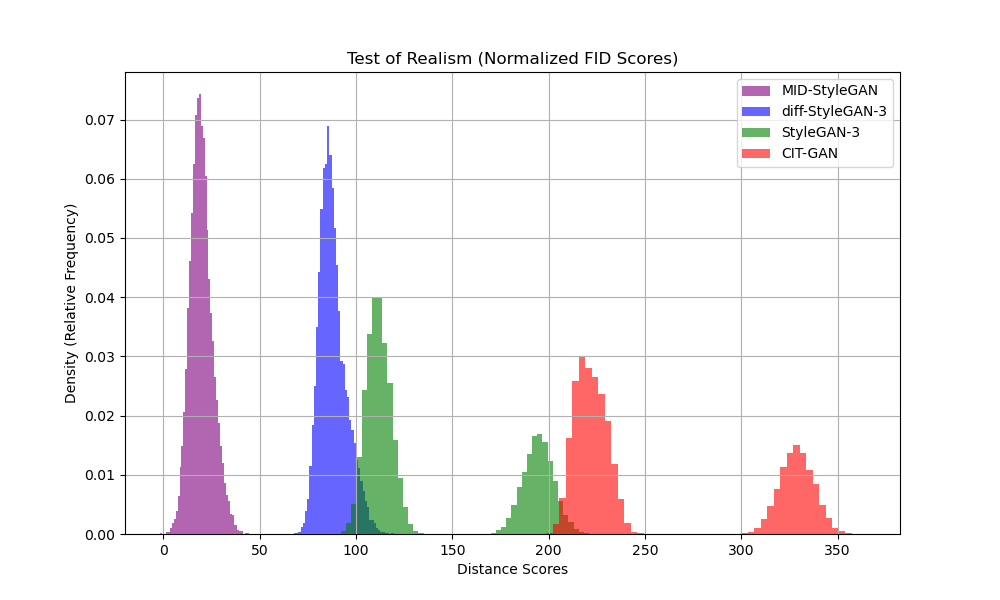}
        \caption{This histogram shows the FID scores of generated images using four generative methods: Proposed method MID-StyleGAN, diff-StyleGAN-3, StyleGAN-3, and CIT-GAN. MID-StyleGAN achieves the lowest FID scores, indicating better image quality, while CIT-GAN and StyleGAN-3 show higher FID scores and inconsistencies, especially across domains like printed eyes.}
        \label{fig:all_histogram}
    \end{subfigure}
    \hfill
    \begin{subfigure}[b]{0.45\textwidth}
        \includegraphics[width=\textwidth]{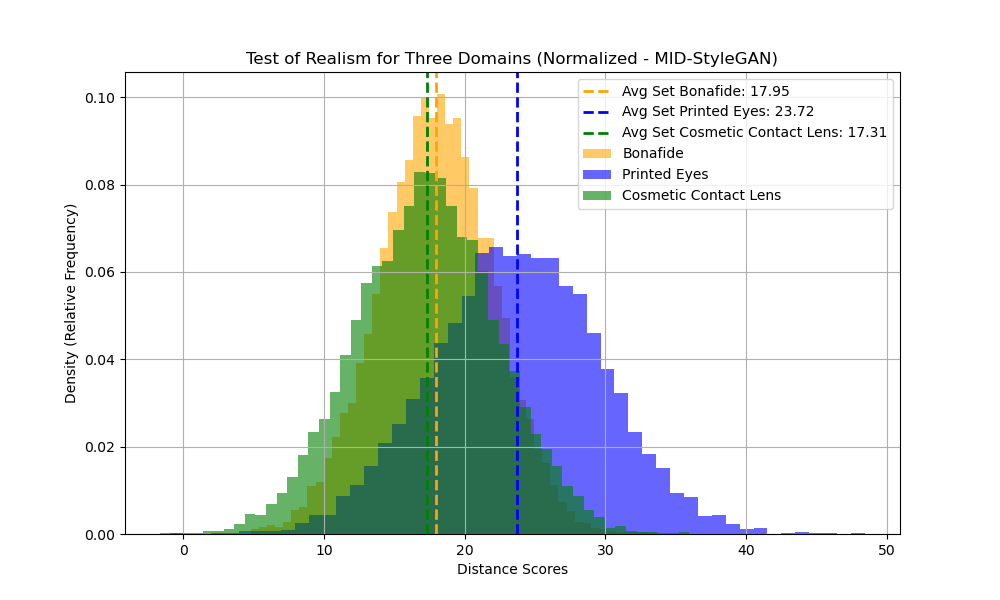}
        \caption{This histogram breaks down MID-StyleGAN's FID scores for the three domains: Bonafide, Printed eyes, and Cosmetic Contact Lens. The Bonafide set has the lowest average FID, showing higher quality, while Printed eyes introduces higher variability and poorer performance.}
        \label{fig:mid_histogram}
    \end{subfigure}
    \caption{Comparison of FID scores across multiple generative methods with respect to proposed method. The first plot shows performance across all methods, while the second focuses on realism of images generated using MID-StyleGAN across different domains.}
    \label{fig:comparison_histograms}
    \vspace{-4mm}
\end{figure}

\noindent
\textbf{Analysis:} The analysis of FID scores reveals that MID-StyleGAN performs best, producing the highest quality images with lower FID scores averaging at 19.71. In contrast, both StyleGAN-3 (average FID of 139.22) and CIT-GAN (average FID of 257.41) exhibit inconsistent performance, reflecting significant variability across domains. The presence of multiple peaks suggests that these other models struggle to maintain consistent quality across different types of synthetic data, especially for printed eyes (Figure \ref{fig:comparison_histograms}).

\begin{figure}
    \centering
    \begin{subfigure}[b]{\linewidth}
        \centering
        \includegraphics[width=0.75\linewidth]{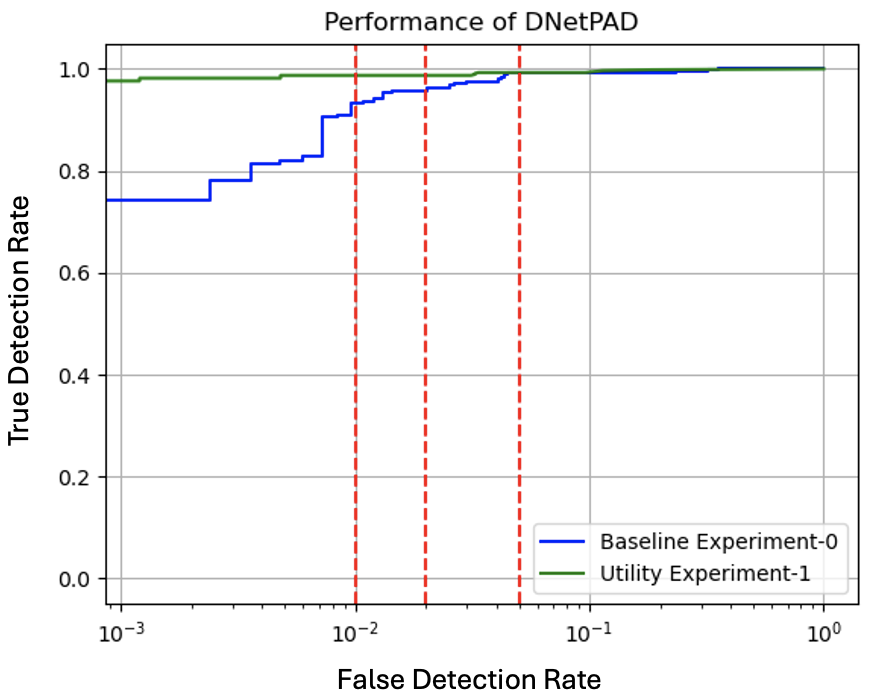}
        \caption{Performance of DNetPAD in baseline Experiment-0 when it is trained using real images from the LivDet-2017 train dataset compared with utility Experiment-1 when it is trained using real and synthetically generated images. The testing is done on the test set of Berc-iris-fake.}
        \label{fig:roc_subfig1}
    \end{subfigure}

    \vspace{1em}  

    \begin{subfigure}[b]{\linewidth}
        \centering
        \includegraphics[width=0.75\linewidth]{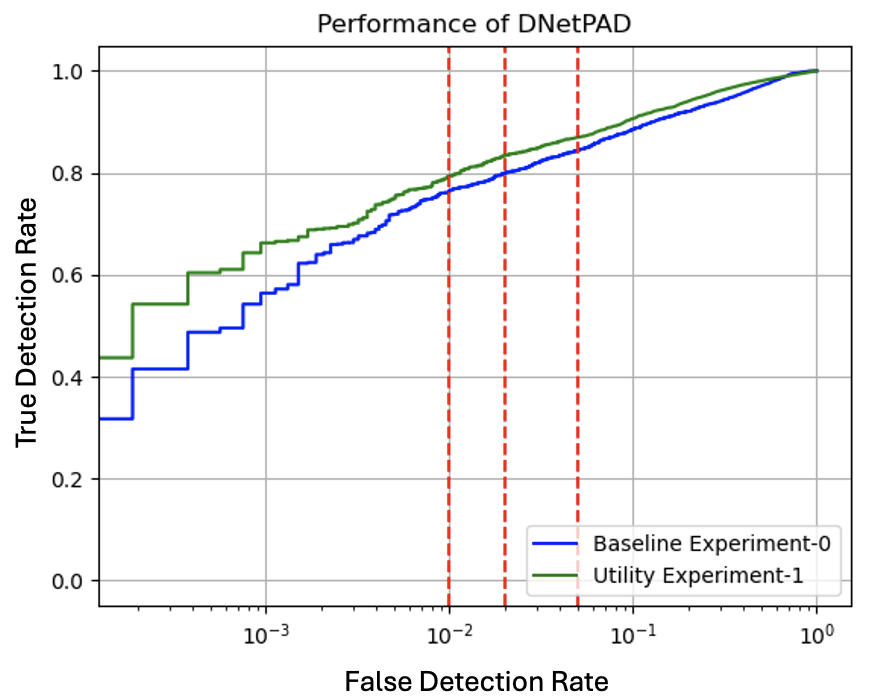}
        \caption{Performance of MobileNet-v2 in baseline Experiment-0 when it is trained using real images from LivDet-2017 train dataset compared with utility Experiment-1 when it is trained using real and synthetically generated images. The testing is done on the test set of LivDet-2020.}
        \label{fig:roc_subfig2}
    \end{subfigure}

    \caption{Performance of iris PA detectors when trained using only real images and also when trained using real+synthetic images showcasing the usefulness of the generated ocular PA dataset.}
    \vspace{-4mm}
    \label{fig:ROCs}
\end{figure}

\subsection{Utility of Generated Dataset}
\label{sec:expUtility}
In this section, we describe experiments to evaluate the utility of the synthetically generated images in training different deep learning based iris PA detection methods, viz., \cite{sharma2021}, VGG-19 \cite{simonyan2014}, ResNet-101 \cite{he2016}, MobileNet-v2 \cite{sandler2018}, AlexNet \cite{krizhevsky2012} and D-NetPAD \cite{sharma20}.

The experiments in this section are done in a cross-dataset scenario, i.e., if the PA detectors are trained on train set of D1, then they are tested on test sets from D3 and D4. D1: Berc-iris-fake \cite{Lee2007} has a total of 2,778 bonafide and 1,820 PAs that is divided using a 70-30 split on each domain (bonafide, printed eyes and cosmetic contact lens), i.e., the train set has 1,944 bonafide and 1,274 PA images while the test set has 834 bonafide and 546 PA images. For D3: LivDet-2017 \cite{Yambay2017} the train-test partition is already provided with 6,563 bonafides and 9,137 PA images in the train set and 5,511 bonafides and 9,356 PA images in the test set. Only the test set for D4: LivDet-2020 \cite{das2020} dataset is available, which has 5,330 bonafides and 6,007 PA images (excluding the post-mortem iris images in this dataset that is not the focus of our study). Note that since the proposed method is an image translative generative method, we set aside D2: CASIA-iris-fake \cite{Sun2014} to be used for synthetic image generation with domain transfer. This ensures that the generated images have no overlap with any images in the test sets. This dataset has a total of 6,000 bonafide images and 1,780 PA images.

\subsubsection{Baseline Experiment-0}
In this experiment, we establish baselines for PA detectors using real bonafide and PA images from the datasets. Training and testing follow a cross-dataset setup: if trained on D1’s train set, testing is done on D3 and D4; if trained on D3, testing is on D1 and D4.

\subsubsection{Utility Experiment-1}
This experiment evaluates the impact of synthetic data on improving PA detection. Unlike the baseline, detectors are trained on both real and synthetic datasets. MID-StyleGAN, trained on D2 (CASIA-iris-fake), generates 7,780 images per domain (bonafide, printed eyes, cosmetic contact lens). These synthetic images are used to augment the training set.

\vspace{2mm} 

\noindent
\textbf{Analysis:} After evaluating the performance of various iris PA detectors on different datasets, we analyzed how the number of samples across domains and variations in the training set can affect the performance of the detectors. This is very clear when comparing the baseline performance of the detectors when trained with D1: Berc-iris-fake dataset which is a comparatively smaller dataset (as shown in Table \ref{tab:pa_comparison1}) versus when trained using the LivDet-2017 dataset (as shown in Table \ref{tab:pa_comparison3}). As seen from the Tables, the VGG-19 detector trained using D1 obtained a TDR of 27.40\% at 1\% FDR when tested on D4: LivDet-2020 dataset, while it obtained 73.08\% TDR at 1\% FDR on D4 when trained using D3: LivDet2017 (which has more number of samples and variations in the dataset). Similar behavior was noticed for the other PA detectors. 

Another evidence of the effect of (a) number of samples across domains and (b) variations in train set on the performance of PA detectors is obtained by augmenting the train set using synthetic ocular samples to introduce more samples per domain with intra-domain variations. Comparing the performance of PA detectors in Table \ref{tab:pa_comparison1} with \ref{tab:pa_comparison2} and Table \ref{tab:pa_comparison3} with \ref{tab:pa_comparison4}, it can be clearly seen that performance of detectors improve after augmenting the train set using synthetic samples. For example, in Table \ref{tab:pa_comparison3} and \ref{tab:pa_comparison4} the performance of DNetPAD \cite{sharma20} when tested on D1 improves from 93.41\% TDR at 1\% FDR to 98.72\% TDR at 1\% FDR. Similar behaviour was seen for other detectors as well.

\subsection{Ablation Study}
To further evaluate MID-StyleGAN, we performed an ablation study by systematically removing its key components to assess their impact on the quality of synthetic ocular images and PAD performance. For all the experiments in this section, we have utilized the same protocol as mentioned in Section \ref{sec:expUtility}.

\subsubsection{Studying Components of Proposed Method}

\vspace{2mm}
\noindent
\textbf{Effect of Style Mixing Regularization:}
Style mixing regularization plays a vital role in encouraging diversity in the generated images by mixing styles from different layers. To assess its impact, we conducted experiments with and without this regularization. When style mixing was removed, we observed a noticeable drop in image quality and diversity. The generated images tended to lack variability across different domains, which hindered their utility for cross-domain analysis in presentation attack detection. The average FID score increased (worsened) by approximately 9.79\%, indicating degraded image quality.

\vspace{2mm}
\noindent
\textbf{Impact of Path Length Regularization: }
Path length regularization ensures smoother transitions in the latent space and improves the consistency of generated images. We performed experiments by disabling this regularization. The results showed that, without path length regularization, the generator produced less consistent outputs, with occasional abrupt changes in image features. The average FID score worsened by approximately 8.12\%, and visual inspection of the generated images revealed artifacts that negatively impacted their realism. This regularization was particularly critical for maintaining the smooth transitions between different ocular domains.

\vspace{2mm}
\noindent
\textbf{Role of Domain-Specific Discriminator: }
The multi-domain discriminator in MID-StyleGAN was specifically designed to handle domain transfer by discriminating images based on their target domain. We conducted an experiment by replacing the multi-domain discriminator with a standard single-domain discriminator. Without the multi-domain capability, the model struggled to enforce domain-specific characteristics in the generated images. The generated PA samples lacked clear domain-specific features, and domain confusion was evident. The average FID score worsened by 20.70\%, suggesting a significant decrease in the quality of the generated domain-transferred images.

\vspace{2mm}
\noindent
\textbf{Effect of Content Preservation Loss: }
We further analyzed the effect of content preservation loss (reconstruction loss) by removing it from the objective function. In this setting, the model generated images that diverged significantly from the input samples, with important features being lost during the domain transfer process. This loss function is crucial for ensuring that key ocular features are retained, even when the domain is altered. Without this component, the model's capacity for realistic and recognizable presentation attack generation was severely compromised.

\vspace{2mm}

Therefore, each component of MID-StyleGAN plays a vital role. Style mixing and path length regularization improve diversity and smoothness, while the multi-domain discriminator ensures domain-specific generation. Content preservation loss retains key ocular features. The proposed architecture balances these elements, producing high-quality domain-transferred images with low FID scores across ablation settings. We also evaluated identity preservation in the generated images by comparing their similarity with source images using the commercial VeriEye matcher.\footnote{\url{www.neurotechnology.com/verieye.html}} Using this matcher, we observed an average similarity score of 471 indicating identity preservation from the source to the generated images.

\subsubsection{Successive Training Using Synthetic Data}
We conducted an experiment where MID-StyleGAN was first trained on {\em real} data, then retrained from scratch in successive generations using only the {\em synthetic} data from the previous version \cite{guo2023, bertrand2024}. The aim was to examine how successive synthetic-only training impacts image quality and PAD performance, both on seen and unseen domains. The first model (Synthetic-1) was trained on real data to generate synthetic images across three domains: bonafide, printed eyes, and cosmetic contact lenses. Each new generation (Synthetic-2 to Synthetic-5) was trained on the preceding synthetic set. At each stage, Fréchet Inception Distance (FID) was computed, and PAD methods were trained on Real + Synthetic data following the protocol in Section~\ref{sec:exp}.

\vspace{2mm}
\noindent
Analysis on Synthetic-1:
Synthetic-1 images were generated using MID-StyleGAN trained on real data. Synthetic-1 images achieved an average FID score of 19.71 (30,000 images; 10,000 per domain), indicating high realism. PAD methods trained on Real + Synthetic-1 showed improved performance compared to using only real data. For instance, DNetPAD’s TDR on D1 improved from 93.41\% to 98.72\% at 1\% FDR (Tables \ref{tab:pa_comparison3} and \ref{tab:pa_comparison4}). Additionally, when DNetPAD was trained using Real D3 + Synthetic-1 without images of cosmetic contact lens (for unseen PA detection), it was observed that the performance dropped from 98.73\% to 85.10\% when tested on D4. 

\vspace{2mm}
\noindent
Analysis on Synthetic-2:
Synthetic-2 images were generated by training MID-StyleGAN on Synthetic-1 images. This dataset yielded a higher (poorer) FID score of 20.36. When Real + Synthetic-2 was used to train PAD methods, slight improvements were observed. On the D4 test set, TDR improved by 0.23\% for VGG-19 and 0.18\% for DNetPAD compared to using Synthetic-1 + Real. Additionally, when DNetPAD is trained using on Real D3 + Synthetic-2 without images of cosmetic contact lens (for unseen PA detection), it was observed that the performance dropped to 86.38\% at 1\% FDR when tested on D4.

\vspace{2mm}
\noindent
Analysis on Synthetic-3:
Synthetic-3 images were generated by training MID-StyleGAN on Synthetic-2 images. This set had an even higher (poorer) FID score of 31.64. Interestingly, when Synthetic-3 + Real was used to train PAD methods, their performance improved further: on the D4 test set, TDR increased by 2.11\% for VGG-19 and 3.42\% for DNetPAD compared to the Synthetic-1 + Real baseline. Additionally, when DNetPAD was trained using Real D3 + Synthetic-3 without images of cosmetic contact lens (for unseen PA detection), it was observed that the performance dropped to 85.98\% at 1\% FDR when tested on D4.

\vspace{2mm}
\noindent
Analysis on Synthetic-4:
Synthetic-4 images were generated by training MID-StyleGAN on Synthetic-3 images. This dataset had a higher (poorer) FID score of 49.25. When Synthetic-4 + Real was used for training, PAD performance started to decline. On D4, TDR dropped by 2.21\% (VGG-19) and 1.43\% (DNetPAD) relative to the Synthetic-1 + Real baseline. Additionally, when DNetPAD was trained using Real D3 + Synthetic-4 without images of cosmetic contact lens (for unseen PA detection), it was observed that the performance dropped to 78.93\% at 1\% FDR when tested on D4.

\vspace{2mm}
\noindent
Analysis on Synthetic-5:
Synthetic-5 images were generated by training MID-StyleGAN on Synthetic-4 images.  Synthetic-5 had the highest (poorest) FID score of 80.74. The use of Synthetic-5 + Real for training, further degraded PAD performance. On D4, TDR decreased by 4.69\% (VGG-19) and 3.02\% (DNetPAD) compared to training with the Synthetic-1 + Real baseline. Additionally, when DNetPAD was trained on Real D3 + Synthetic-5 without the images from cosmetic contact lens (for unseen PA detection), it was observed that the performance dropped to 74.91\% at 1\% FDR when tested on D4.

\vspace{2mm}
\noindent
Overall, early generations such as Synthetic-2 maintain relatively low FID scores and boost PAD performance on both seen and unseen domains, indicating that synthetic data can enhance domain generalization when used for augmentation. However, later generations (Synthetic-4, Synthetic-5) show increased FID, reduced realism, and drops in both seen- and unseen-domain performance, likely due to compounding noise from repeated synthetic-only training \cite{guo2023}. Synthetic augmentation is thus most effective in early cycles, especially for improving robustness to unseen attack types.


\section{Conclusion and Future Work}
\label{sec:conclusion}
The proposed approach for multi-domain image translation, which combines a GAN with a diffusion model, within the context of iris presentation attack detection, effectively ensures that the generated ocular images pertain to a specified target domain. By leveraging the domain classification loss, the model is trained to produce images that not only exhibit realistic features but also align well with the desired domain; therefore, they can then be used to train a more accurate and robust PA detector. At present, our approach does not specifically aim to generate entirely new identities. This decision is based on the nature of the presentation attack detection task, where the primary concern is distinguishing between bonafide and attack images rather than deducing identities. Consequently, the model may replicate certain identity features from the training data, which is acceptable within the context of this specific application. However, we recognize the importance of privacy considerations in synthetic image generation. Moving forward, our goal is to refine this approach to be more privacy-conscious by ensuring that the generated images do not replicate identity characteristics from the training data.
{
    \small
    \balance
    \bibliographystyle{ieeenat_fullname}
    \bibliography{main}
}

\end{document}